\documentclass[10pt]{article}
\usepackage[margin=1in]{geometry}
\usepackage{times}
\usepackage{graphicx}
\usepackage{booktabs}

\usepackage{multirow}
\usepackage{xcolor}
\usepackage{amsmath}
\usepackage{amssymb}
\usepackage{hyperref}
\usepackage{hyperref}
\hypersetup{draft}
\hypersetup{
    colorlinks=false,
    pdfborder={0 0 0},
    hidelinks
}
\usepackage{float}
\usepackage{subcaption}
\usepackage{cuted}
\usepackage{listings}
\usepackage{arydshln}
\usepackage{setspace}
\usepackage{titlesec}

\usepackage[numbers,sort&compress]{natbib}

\makeatletter
\renewcommand\@cite[2]{\textsuperscript{#1}}
\renewcommand\cite[1]{\textsuperscript{\citenum{#1}}}
\makeatother

\makeatletter
\renewcommand{\paragraph}{%
  \@startsection{paragraph}{4}{\z@}%
  {1ex}{-0.5ex}%
  {\normalfont\normalsize\bfseries}%
}
\makeatother

\titleformat{\section}
  {\normalfont\bfseries\fontsize{10}{15}\selectfont}{\thesection}{1em}{}
\titleformat{\subsection}
  {\normalfont\bfseries\fontsize{10}{13}\selectfont}{\thesubsection}{1em}{}

\titlespacing*{\section}{0pt}{0.7ex}{0.4ex}
\titlespacing*{\subsection}{0pt}{0.5ex}{0.3ex}

\title{\bfseries RadAnnotate: Large Language Models for Efficient and Reliable Radiology Report Annotation}
\author{
\textbf{Saisha Pradeep Shetty, MSc}$^{1}$,
\textbf{Roger Eric Goldman, MD, PhD}$^{2}$,
\textbf{Vladimir Filkov, PhD}$^{1}$
\\
$^{1}$Department of Computer Science, University of California, Davis, CA, USA \\
$^{2}$Department of Radiology, University of California, Davis, CA, USA \\
}
\date{}

\begin{document}
\maketitle
\vspace{-1em}

\begin{abstract}
\textit{
Radiology report annotation is essential for clinical NLP, yet manual labeling is slow and costly. We present RadAnnotate, an LLM-based framework that studies retrieval-augmented synthetic reports and confidence-based selective automation to reduce expert effort for labeling in RadGraph. We study RadGraph-style entity labeling (graph nodes) and leave relation extraction (edges) to future work. First, we train entity-specific classifiers on gold-standard reports and characterize their strengths and failure modes across anatomy and observation categories, with uncertain observations hardest to learn. Second, we generate RAG-guided synthetic reports and show that synthetic-only models remain within 1–2 F1 points of gold-trained models, and that synthetic augmentation is especially helpful for uncertain observations in a low-resource setting, improving F1 from 0.61 to 0.70. Finally, by learning entity-specific confidence thresholds, RadAnnotate can automatically annotate 55–90\% of reports at 0.86–0.92 entity match score while routing low-confidence cases for expert review.}
\end{abstract}

\section{Introduction}
The development of AI-assisted clinical tools is constrained by the scarcity of high-quality annotated medical data. Even small inconsistencies in entity boundaries or negation cues can propagate into errors in critical downstream tasks like epidemiologic surveillance and decision-support modeling~\cite{ostropolets2024scalable,huang2024critical,andrade2024boundary}.
However, producing such labels is slow and costly: radiologists must interpret subtle findings, handle uncertainty and negation in free text, and maintain consistency across nuanced clinical phrasing, making large-scale expert annotation difficult to sustain.
Annotating RadGraph-style~\cite{jain2021radgraph} entity and relation labels takes approximately 1--3 minutes per chest radiograph report, so a 50,000-report corpus requires 800--2,500 expert hours, creating a substantial barrier for institutions seeking to build structured clinical NLP resources. RadGraph-style denotes the entity–relation information extraction schema introduced in the RadGraph dataset, which structures radiology reports into anatomy/observation entities and directed relations (e.g., located\_at, modify, suggestive\_of). In practice, RadGraph is used as a standard benchmark and training target for report-level entity/relation extraction models, and the resulting structured graphs support downstream use cases such as structured report retrieval/cohort construction and multimodal learning with paired chest radiographs. Prior work in radiology NLP spans rule-based systems, clinical NLP pipelines, and supervised transformer-based extractors (e.g., RadGraph Benchmark / DyGIE++; NER micro-F1  0.94/0.91 on MIMIC-CXR/CheXpert test sets), but these efforts primarily optimize extraction accuracy, whereas our goal is to characterize LLM-based annotation and enable confidence-guided selective automation. Recent evaluations also show that generative LLMs can perform zero-shot extraction in clinical text~\cite{dorfner2024comparing,abdullah2025automated,hu2024zero}, though reliability varies across scenarios~\cite{reichenpfader2024scoping}.

Despite this progress, two deployment-oriented gaps remain. First, although several studies generate synthetic clinical text~\cite{hernandez2022synthetic,suvalov2025using,tang2023synthetic,liu2024generating}, none evaluate whether retrieval-augmented synthetic radiology reports can serve as reliable training data for RadGraph-style entity annotation, especially for sparse categories such as uncertain observations (OBS-U), which have very few gold examples. Second, existing automated annotation systems~\cite{goel2023llms,shin2025twostage,altalla2025radiology} do not incorporate confidence-based mechanisms to determine when predictions are safe to automate versus when expert review is needed.

Prior work addresses parts of these challenges but not the specific problems we study. 
Goel et al.~\cite{goel2023llms} examine LLM-assisted information extraction but do not evaluate synthetic training data for structured annotation. 
Shin et al.~\cite{shin2025twostage} present a two-stage entity--relation pipeline tailored to narrow lesion-location or diagnosis-episode pairs, without mechanisms for assessing prediction reliability. 
Altalla et al.~\cite{altalla2025radiology} generate radiology text but do not produce RadGraph-style structured labels or evaluate annotation quality.  
To our knowledge, no prior work evaluates (1) RAG-guided synthetic radiology reports as training data for RadGraph-style entity annotation or (2) entity-specific confidence thresholds for selective automation of RadGraph labels.
In this work, we introduce \textit{RadAnnotate}, a framework that integrates entity-specific LLM classifiers, retrieval-augmented synthetic data generation, and confidence-guided selective automation. RadAnnotate evaluates whether LLM-based annotation can safely reduce manual workload by automatically handling high-confidence cases while routing ambiguous reports for expert review.

Our contributions are threefold:
(1) an evaluation of LLM-based radiology report annotation using RadGraph~\cite{jain2021radgraph}, characterizing strengths and limitations across entity types;
(2) a RAG-based pipeline for generating synthetic reports and assessing their utility under varying data regimes, including extreme low-resource settings; and
(3) a confidence-based selective automation framework enabling principled trade-offs between automation coverage and manual review.

RQ1–RQ3 provide the structure through which we realize the above contributions. Specifically, RQ1 establishes the empirical foundation for evaluating LLM-based RadGraph-style annotation, RQ2 stress-tests our approach under data scarcity by assessing the value of retrieval-augmented synthetic reports, and RQ3 translates these findings into a deployment-oriented selective automation framework by quantifying safe automation coverage and expert time savings. Our research addresses three deployment-centered questions:
\begin{enumerate}

\item \textbf{RQ1:} Can a reliable entity annotator be built using gold data from a medium-sized RadGraph-style corpus?
\item \textbf{RQ2:} When rare categories (e.g., OBS-U) have few labeled examples, can retrieval-augmented synthetic reports improve performance?
\item \textbf{RQ3:} Using model confidence, how many reports can be safely auto-annotated, and how much expert time can realistically be saved?
\end{enumerate}


\section{Methodology}

\textit{Data and Annotation Schema:}
We evaluated our approach on the RadGraph dataset \cite{jain2021radgraph}, which provides entity- and relation-level annotations for chest radiology reports. We focus on four entity types defined in its schema: ANAT-DP (anatomy, definitely present), OBS-DP (observation, definitely present), OBS-DA (observation, definitely absent), and OBS-U (observation uncertain). These four entity types follow directly from the RadGraph schema design, which separates anatomical mentions from imaging observations and further subdivides observations into three clinically meaningful uncertainty states (definitely present, uncertain, and definitely absent), yielding one anatomy category and three observation categories. Examples include “lungs’’ for ANAT-DP, “effusion’’ for OBS-DP, “no consolidation’’ for OBS-DA, and “possible edema’’ for OBS-U. Each token span receives exactly one label, with negation or uncertainty cues (e.g., “no,” “possible’’) absorbed into the entity type rather than annotated separately. We evaluate annotation quality using standard entity-level precision, recall, and F1 scores, comparing each predicted clinical entity with its corresponding gold-standard label from RadGraph. Following annotation, these entities are used to train entity-specific classifiers, allowing the system to learn both positive and negative contexts for each category.

\textit{Sentence-Level Data Processing:} The original RadGraph dataset contains 425 training and 75 development reports from MIMIC-CXR, plus a 100-report test set split evenly between MIMIC-CXR and CheXpert. To maximize the utilization of training data from this limited corpus, we implement a sentence-level splitting strategy. Each radiology report, typically containing 4–5 sentences, is segmented into individual sentences, each sentence treated as an independent training example while preserving its original entity annotations. This approach is valid because each sentence in radiology reports typically contains complete clinical observations that can be independently analyzed. The sentence-splitting strategy expands our training corpus from 425 reports to approximately 2,425 sentence-level instances,
with development and test sets expanding to 480 and 511 instances respectively, providing sufficient data density for robust model training.

\textbf{\textit{Entity-Specific Model Training:}}
\label{sec:dist} We train four separate \emph{Qwen2.5-7B} models, one for each RadGraph entity type (ANAT-DP, OBS-DP, OBS-DA, OBS-U). Rather than building a single multi-class model, we adopt this entity-specific approach because the label distribution is highly imbalanced: in our corpus, ANAT-DP appears in 2{,}033 sentences and OBS-DP in 1{,}856, whereas OBS-DA and OBS-U occur in only 552 and 309 sentences, respectively, mostly in narrow clinical contexts involving explicit negation or uncertainty. A unified model would therefore be dominated by high-frequency labels, causing it to overfit common entities and systematically underperform on rare but clinically important categories.
Each entity-specific model is trained as an instruction-following \emph{generative extractor} that outputs a structured JSON list of spans for its target entity type. 

We fine-tune \emph{Qwen2.5-7B} using supervised instruction tuning with QLoRA, where each training example contains an entity-specific instruction, the input report text, and a JSON output specifying \texttt{entity\_type}, \texttt{entity\_value}, and character-level \texttt{start\_position}/\texttt{end\_position}. We use a fixed prompt template (“Task / Input / Output”) across entities, varying only the target-label instruction. For training and evaluation, we construct sentence-level examples in this instruction-tuning format. Sentences containing at least one span of the target entity type are treated as positives, and the target output is the JSON list of extracted spans. Sentences that do not contain the target entity type, even if they include other entities, are paired with an empty list (\texttt{[]}) as the output. This yields four per-entity datasets whose positive-to-negative ratios reflect the natural prevalence of each entity type, with substantially fewer positives for rare categories such as OBS-DA and OBS-U. We parse the generated JSON and align predicted spans back to the input text using the provided character offsets for evaluation.

\textbf{\textit{RAG-Enhanced Synthetic Data Generation:}} To check whether RAG-enhanced synthetic data can achieve comparable performance to expert annotation, we develop a multistage RAG-enhanced synthetic data generation framework (Figure ~\ref{fig:rag_pipeline}). Our approach combines frequency-based medical keyword extraction with retrieval-augmented generation to produce clinically realistic synthetic reports, followed by LLM-based quality validation to ensure annotation accuracy. The framework consists of three main stages:

\begin{figure}[htbp]
\centering
\includegraphics[width=0.65\textwidth]{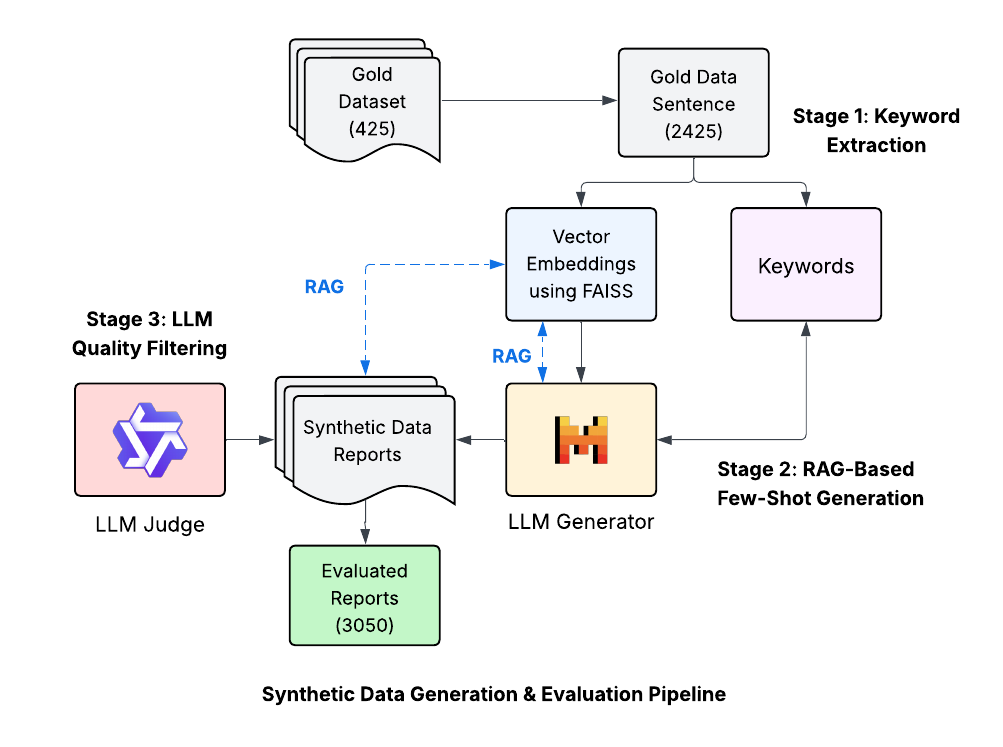}
\caption{RAG-Enhanced Synthetic Data Generation Pipeline}
\label{fig:rag_pipeline}
\end{figure}

\textit{Frequency-Based Medical Keyword Extraction:} We implement a frequency analysis approach to extract the most salient medical keywords from our 2,425 sentence training corpus. Keywords are ranked by frequency of occurrence, prioritizing clinical terminology including anatomical references (e.g., "lung," "heart"), pathological findings (e.g., "pneumonia," "effusion"), and diagnostic descriptors (e.g., "enlarged," "normal"). We select high-frequency terms to serve as semantic queries for the subsequent retrieval-augmented generation stage. This frequency-based selection ensures that synthetic reports focus on the most common and clinically relevant terminology found in real radiology data.

\textit{RAG-Based Few-Shot Generation:}
To generate high-quality synthetic radiology reports, we employ a Retrieval-Augmented Generation (RAG) pipeline. We use extracted medical keywords to retrieve semantically similar expert-annotated reports, which serve as few-shot examples\cite{brown2020language} to guide the generation of synthetic reports\cite{gao2023retrieval}.

\textit{Semantic Retrieval:} All reference reports are embedded using the \textit{all-MiniLM-L6-v2} model\cite{reimers2019sentence} and indexed using FAISS\cite{johnson2019billion}. During generation, we construct query embeddings from selected medical keywords and retrieve the top-k most semantically similar reports (k = 2 or 3) to use as few-shot examples.

\textit{Structured Prompt Design:} We employ a structured prompt that instructs the generation model (\textit{Mixtral-8x7B-Instruct-v0.1}, accessed via Mistral Inference) to simultaneously generate synthetic radiology sentences and their corresponding entity annotations. The model is guided to generate clinically plausible one-sentence radiology reports with appropriate entity labels, following the patterns demonstrated in the retrieved few-shot examples\cite{lewis2020retrieval}. This approach ensures that synthetic reports reflect the structure, terminology, and entity distributions found in real radiology data.

\textit{Diversity Control and Validation Readiness:} To ensure diversity in synthetic data generation, we randomize keyword selection across the pool of extracted clinical terms and generate reports in batches to minimize repetition. This retrieval-based generation process produces synthetic reports with corresponding token-level entity labels ready for model training.

\textit{LLM Quality Filtering:}
We implement a large language model (LLM)–based quality control mechanism, referred to as the LLM Judge, using \emph{Qwen2.5-32B}. For each synthetic report, the Judge receives the generated radiology text, its associated entity labels, and the top three most similar gold-standard reports retrieved as reference examples. Given this context, the Judge evaluates whether the synthetic annotations are accurate and clinically consistent. The Judge is instructed to: (1) validate each label against the report, (2) remove stop words or invalid labels, (3) correct mislabels when the context indicates uncertainty or negation, (4) prevent hallucination by avoiding the introduction of new terms, and (5) add missing labels when corresponding terms appear in the text. The output is a validated JSON mapping of words to corrected entity labels. We also remove synthetic reports that exactly match a gold-standard report to preserve data integrity, resulting in a final synthetic dataset of 3,050 reports.

\textbf{Evaluation for Synthetic Data Assessment:} To evaluate the role of synthetic data in the radiology annotation process, we train four additional entity-specific models using the synthetic reports generated by our framework. These models mirror the gold-data setup, enabling controlled comparisons. Our evaluation consists of three experiments: 
\begin{enumerate}
    \item \textbf{Experiment 1}: Compares gold-only and synthetic-only models trained under identical conditions to assess whether synthetic data can reproduce gold-level performance
    \item \textbf{Experiment 2}: Tests whether adding synthetic positives to the full gold dataset improves performance for any entity type
    \item \textbf{Experiment 3}: Analyzes a low-resource setting by training on a small gold seed (50 reports) and incrementally adding synthetic examples to measure how performance scales. 
\end{enumerate}
This design allows us to assess both the fidelity of synthetic data and its usefulness under varying data regimes.

\textit{Distributional Similarity Check:}
We assess whether synthetic reports resemble real radiology text by comparing their embedding distributions. Internal coherence is measured as the average similarity of each report to its nearest neighbors within the same corpus, indicating how linguistically consistent (or noisy) each dataset is. Cross-set similarity measures how closely synthetic reports match real ones. We also generate a t-SNE projection to visually confirm whether real and synthetic embeddings overlap i.e., whether they occupy the same semantic space rather than forming separate clusters. In addition, we use the cross-set similarity distribution as a near-duplication check and inspect for unusually high-similarity outliers that could indicate leakage from the gold corpus. Although RadGraph is de-identified, this analysis helps reduce the risk of inadvertent memorization or evaluation contamination.

\begin{figure*}[h]
\centering
\includegraphics[width=\textwidth]{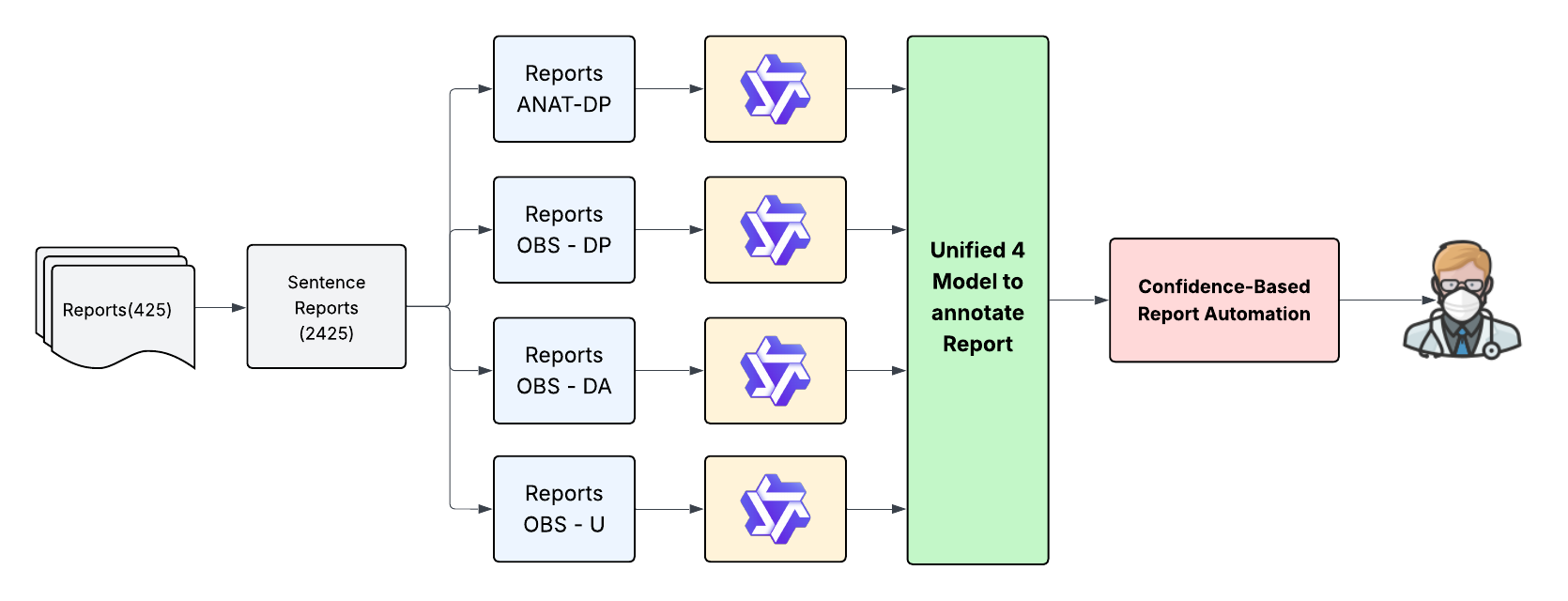}
\caption{Overview of the multi-model pipeline used for confidence-based automation.}
\label{fig:multimodel-pipeline}
\end{figure*}

\textbf{Automation Model Training:} For the automation stage, each entity-specific classifier is trained independently using only gold-standard data. We include explicit negative examples by retaining all sentences and treating those without the target entity (even if other entity types appear) as negatives with an empty label for that entity, while sentences containing the target entity are positives. This enables each classifier to learn a clear distinction between the presence and absence of its assigned entity type. The positive-to-negative ratio follows the natural distribution of each entity type, allowing high-frequency entities such as ANAT-DP and OBS-DP to be learned from diverse contexts, while low-frequency entities such as OBS-DA and OBS-U require proportionally more negatives. An overview of the full multi-model pipeline is shown in Figure~\ref{fig:multimodel-pipeline}.

\textit{Confidence Scoring and Threshold Discovery:}
Each classifier outputs a logit for the target entity, which is converted to a confidence score \cite{geng2024survey} using a sigmoid transformation:
\[
\hat{p} = \frac{1}{1 + e^{-\mathrm{logit}}}.
\]
To identify confidence cutoffs, we use the 511-sentence-level held-out test set, which is split evenly into a validation split for threshold selection and a test split for final automation evaluation. On the validation set, thresholds from 0.00 to 1.00 are swept in increments of 0.01. At each threshold, we compute the entity match score and coverage, where coverage is the fraction of sentences whose predicted entities meet or exceed the confidence cutoff. We accept a sentence for automation only if all predicted entities pass their respective thresholds; otherwise, the sentence is deferred for manual review. We compute an Entity Match Score, which compares the set of extracted entities in a sentence to the gold entity set using exact span matches:
\[
\mathrm{Entity\ Match\ Score} = 
\frac{\mathrm{TP}}{\mathrm{TP} + \mathrm{FP} + \mathrm{FN}}.
\]
Here, TP counts predicted entities that exactly match a gold entity, FP counts extra predicted entities not in gold, and FN counts gold entities that were missed. We omit true negatives (TN) because the set of non-entity spans is extremely large in extraction tasks, making TN-based score uninformative and inflated. 

For each entity type, we record the lowest threshold achieving target match scores of 0.80, 0.85, 0.90, and 0.95, producing four threshold curves and four sets of recommended cutoffs. If an entity type does not reach a specified target match score during threshold discovery, we assign the threshold that attains the highest match score available for that entity type as a fallback. This ensures that every entity has a defined confidence cutoff for downstream automation.

\textit{Report-Level Automation and Conflict Resolution:}
Although thresholds are selected independently per entity type, automation is evaluated at the full-report level. All four classifiers are applied to each report, and predictions are merged. Overlapping or conflicting spans are consolidated to ensure that each token span receives a single label. When multiple candidate labels are predicted for the same span, we resolve conflicts by selecting the label with the highest confidence margin \((\hat{p} - \tau)\), where \(\hat{p}\) is the confidence score and \(\tau\) is the threshold for that entity type. A report is accepted for automation only if at least a proportion \(p\) of its predicted entities meet or exceed their thresholds, where
\[
p \in \{0.90,\;0.95\}.
\]
Reports that fail this criterion are routed for manual review. This full-pipeline evaluation, illustrated in Figure~\ref{fig:multimodel-pipeline}, mirrors real deployment and ensures that the automation results reflect the end-to-end system behavior.

\section{Results}
\label{sec:methods}
\subsection{RQ1: Gold Standard Data Performance}

\begin{table}[htbp]
\centering
\caption{Annotation Performance Using Gold-Only Training Data}
\label{tab:ner-gold}
\begin{tabular}{l c c c c c c}
\toprule
\textbf{Entity Type} & \textbf{Train Reports} & \textbf{Test Reports} & \textbf{Precision} & \textbf{Recall} & \textbf{F1} \\
\midrule
ANAT-DP & 2033 & 430 & 0.9302 & 0.9444 & 0.9372 \\
OBS-DP  & 1856 & 386 & 0.9226 & 0.7967 & 0.8550 \\
OBS-DA  & 552  & 143 & 0.9664 & 0.9489 & 0.9576 \\
OBS-U   & 309  & 52  & 0.8276 & 0.6857 & 0.7500 \\
\midrule
\textbf{Aggregate} & 4750 & - & 0.9117 & 0.8439 & 0.8747 \\
\bottomrule
\end{tabular}
\vspace{-1mm}
\end{table}

Table \ref{tab:ner-gold} summarizes the performance of the four entity-specific classifiers trained solely on gold-standard reports. The models perform strongly overall (aggregate F1 = 0.8747), indicating high-quality radiology annotation is achievable with a modest corpus. Because each classifier is evaluated only on test sentences containing its target entity, the effective test size differs across ANAT-DP, OBS-DP, OBS-DA, and OBS-U. ANAT-DP performs best (F1 = 0.9372), while OBS-DP shows lower recall (0.7967), consistent with greater lexical and contextual variability in observation phrases. OBS-DA performs well despite fewer examples (F1 = 0.9576), likely due to regular negation patterns (e.g., “no effusion”), whereas OBS-U is lowest (F1 = 0.7500), reflecting ambiguity and limited coverage of uncertainty cues. 

To provide a non-LLM baseline, we evaluated the released RadGraph DyGIE++ (trained for joint entity and relation extraction) checkpoint on our test reports, using only its entity predictions and aggregating them into our four entity categories. It achieved F1 = 0.890 (vs. 0.8747 for our gold-trained models), showing that our LLM approach performs nearly on par with established state-of-the-art clinical extractors.
These results show that while gold-only training produces robust models, entity characteristics such as frequency and linguistic variability strongly influence performance. This motivates RQ2, where we explore whether synthetic data can help address these limitations, particularly for rare or ambiguous categories like OBS-U. RQ1 evaluated entity extraction under a presence-conditioned setting to isolate span-level extraction behavior, while in RQ3 we will evaluate the merged pipeline on the full test distribution, including negative contexts, to reflect deployment where the model must decide both presence and absence.
\subsection{RQ2: Can Synthetic Data Enhance Annotation Performance?}

First, we explore how closely synthetic reports resemble real radiology text by comparing their embedding distributions. The synthetic corpus shows strong semantic alignment with the real reports (cross-set similarity = 0.877). It exhibits higher internal coherence than real reports (0.914 vs 0.850 for real reports), suggesting linguistically consistent and clinically plausible generation. 
A t-SNE projection of sentence embeddings (Figure ~\ref{fig:sim}) visualizes this overlap, with synthetic and real text occupying closely aligned regions in embedding space. 

\begin{center}
\includegraphics[width=0.5\textwidth]{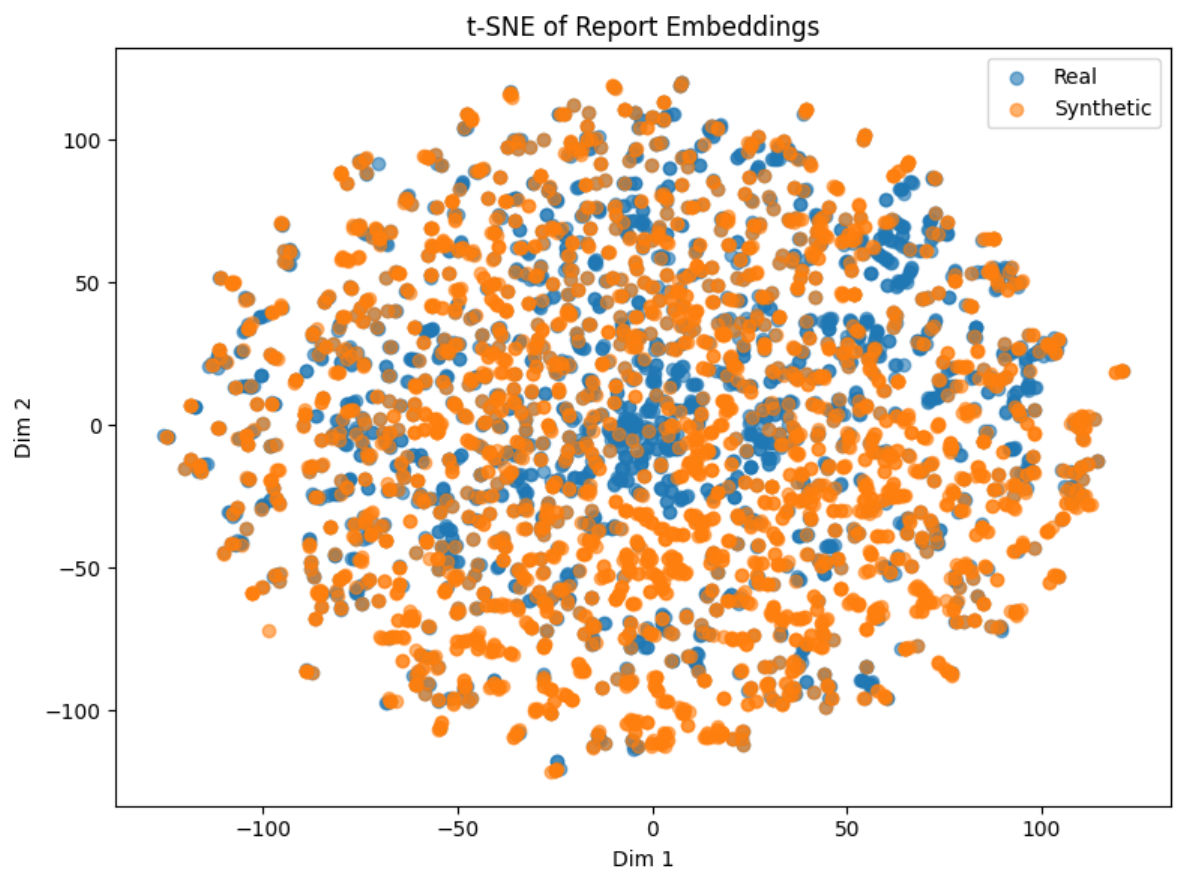}
\captionof{figure}{t-SNE visualization of sentence embeddings shows real and synthetic data overlap significantly.}
\label{fig:sim}
\end{center}

\paragraph{Gold vs.\ Synthetic Annotation Performance:}

To examine whether semantic similarity translates into downstream fidelity, we trained identical \emph{Qwen2.5-7B} classifiers on equal-sized gold and synthetic subsets for each RadGraph entity type. Note that the RQ2 test sets are smaller than those in RQ1 because any report that appeared in the synthetic-generation seed pool was removed from the corresponding gold test split to avoid data leakage. Synthetic-only models achieve performance close to gold-trained models, typically within 1--2 F1 points (Table~\ref{tab:ner-full-combined}). The smallest gaps occur in structured categories such as ANAT-DP (0.9315 to 0.9119) and OBS-DA (0.9539 to 0.9373), where consistent anatomical or negation patterns lead to strong transfer. For OBS-DP, recall remains similar but precision decreases, indicating a mild tendency toward over-labeling when trained on synthetic text. For OBS-U, performance is nearly identical between gold and synthetic models (0.7564 vs.\ 0.7561), suggesting that even limited uncertainty-related seed examples are sufficient to reproduce characteristic expressions such as “possible,” “may represent,” and “could be.”

\begin{table*}[h]
\centering

\caption{Comparison of Annotation Performance Across Gold, Synthetic, and Gold+Synthetic (30\%) Training Data.}
\label{tab:ner-full-combined}
\small

\begin{tabular}{l c c c c c c c}
\toprule
\textbf{Entity Type} & \textbf{Source} & \textbf{Gold Count} & \textbf{Syn Count} & \textbf{Test Reports} & \textbf{Precision} & \textbf{Recall} & \textbf{F1} \\
\midrule

\multirow{3}{*}{ANAT-DP}
    & Gold      & 2033 & 0    & 404 & 0.9242 & 0.9390 & 0.9315 \\
    & Syn       & 0    & 2033 & 404 & 0.9128 & 0.9110 & 0.9119 \\
    & Gold+Syn  & 2033 & 610  & 404 & 0.9152 & 0.9060 & 0.9106 \\
\midrule

\multirow{3}{*}{OBS-DP}
    & Gold      & 1856 & 0    & 367 & 0.9216 & 0.7920 & 0.8519 \\
    & Syn       & 0    & 1856 & 367 & 0.8865 & 0.8010 & 0.8416 \\
    & Gold+Syn  & 1856 & 556  & 367 & 0.9104 & 0.7268 & 0.8083 \\
\midrule

\multirow{3}{*}{OBS-DA}
    & Gold      & 552  & 0    & 129 & 0.9635 & 0.9446 & 0.9539 \\
    & Syn       & 0    & 552  & 129 & 0.9498 & 0.9251 & 0.9373 \\
    & Gold+Syn  & 552  & 157  & 129 & 0.9216 & 0.9186 & 0.9201 \\
\midrule

\multirow{3}{*}{OBS-U}
    & Gold      & 309  & 0    & 41  & 0.8310 & 0.6941 & 0.7564 \\
    & Syn       & 0    & 309  & 41  & 0.7848 & 0.7294 & 0.7561 \\
    & Gold+Syn  & 309  & 93   & 41  & 0.8182 & 0.7412 & 0.7778 \\

\bottomrule
\end{tabular}

\end{table*}
Overall, these results show that RAG-generated synthetic reports capture the main patterns of real radiology text and work well as a low-resource substitute for gold data. Synthetic-only models come very close to gold-trained models for the clearer entity types, while small gaps remain for categories with more variability such as OBS-DP. This motivates the next experiment, where we test whether synthetic samples can also add value when gold data is already available.


\paragraph{Synthetic Data Augmentation Effect:} Across three of the four entity types, adding 30\% synthetic data leads to a small decline in performance. The 30\% ratio provides meaningful augmentation without overwhelming the gold distribution. ANAT-DP, OBS-DP, and OBS-DA  show modest decreases in F1, with OBS-DA having the largest drop (0.9539 to 0.9201). OBS-U shows a slight improvement (0.7564 to 0.7778), suggesting that for rare uncertainty-related findings, the synthetic reports signal can broaden the model’s decision boundary. In general, synthetic augmentation can help with the rare instances and shows only mild degradation when sufficient gold data are available. 

Error analysis: To explain the F1 decreases in Table 2, we checked whether synthetic augmentation introduces annotation noise. We measured unanchorable spans (entity values not present in the report text) and found a rate of 0 across all entity types, suggesting no hallucinated labels. However, synthetic supervision can still be incomplete even when grounded if it omits valid spans that appear in the report. To assess this, we ran a lexicon-coverage check: for each entity type, we built a gold-derived lexicon (top-500 most frequent entity strings) and measured how often lexicon-matched spans in synthetic reports were unlabeled. This indicated notable missing-label rates (ANAT-DP: 22.9\%; OBS-DP: 52.9\%; OBS-DA: 42.8\%; OBS-U: 44.7\%). Since lexicon matching is context-blind and may overestimate omissions, especially for short or generic tokens—we treat this as suggestive rather than definitive. Overall, the F1 drops under augmentation are more consistent with missing labels (incomplete supervision) than with hallucination.

\paragraph{Incremental Synthetic Augmentation Under Low-Resource Conditions:}
Since \textsc{OBS-U} contains the fewest gold reports in our corpus, we intentionally simulate an extreme low-resource condition by restricting the gold seed to only 50 reports. This reflects the practical reality that certain clinical findings appear infrequently, leaving only a small number of high-quality annotations available for training.  To isolate the true effect of synthetic augmentation, all synthetic samples are generated solely from these same 50 reports rather than from the full corpus, and the retrieval pool is restricted to these 50 reports so that no external information enters the low-resource seed. For each synthetic ratio (0\%–150\%), we train five independently initialized models to account for stochastic variation in optimization and data sampling. Averaging across five runs provides a stable estimate of performance and reduces the sensitivity to random seeds present in low-resource settings. For each ratio, we report average precision, recall, and F1 across runs.

\begin{table}[h!]
\centering
\caption{Incremental synthetic augmentation results for OBS-U using 50 gold reports.}
\label{tab:syn-add}
\small
\begin{tabular}{cccccc}
\toprule
\textbf{Gold} & \textbf{Syn} & \textbf{\% Syn} 
& \textbf{Precision} & \textbf{Recall} & \textbf{F1} \\
\midrule
50 & 0  & 0\%   & 0.6170 $\pm$ 0.0597 & 0.5976 $\pm$ 0.0228 & 0.6053 $\pm$ 0.0297 \\
50 & 12 & 25\%  & 0.6378 $\pm$ 0.0617 & 0.6329 $\pm$ 0.0554 & 0.6309 $\pm$ 0.0213 \\
50 & 25 & 50\%  & 0.6953 $\pm$ 0.0449 & 0.6329 $\pm$ 0.0262 & 0.6622 $\pm$ 0.0307 \\
50 & 37 & 75\%  & 0.7272 $\pm$ 0.0261 & 0.6847 $\pm$ 0.0389 & 0.7051 $\pm$ 0.0317 \\
50 & 50 & 100\% & 0.7139 $\pm$ 0.0231 & 0.7082 $\pm$ 0.0417 & 0.7103 $\pm$ 0.0247 \\
50 & 62 & 125\% & 0.7275 $\pm$ 0.0275 & 0.6871 $\pm$ 0.0321 & 0.7060 $\pm$ 0.0212 \\
50 & 75 & 150\% & 0.7262 $\pm$ 0.0734 & 0.6800 $\pm$ 0.0382 & 0.7001 $\pm$ 0.0418 \\
\bottomrule
\end{tabular}
\end{table}

Introducing synthetic data monotonically boosts performance over the 50-report gold baseline, Table~\ref{tab:syn-add},
F1 increases from 0.61 at 0\% synthetic data to 0.63 at 25\%, and to 0.66 at 50\%. 
Performance consistently rises to 0.70 at a 1:1 synthetic-to-gold ratio (100\%), 
beyond which gains begin to stabilize: 125\% and 150\% synthetic data yield F1 scores of 0.706 and 0.700, respectively. These results show that a small gold seed can be effectively augmented with synthetic examples, with the greatest improvements occurring in the 75 to 100\% range, after which the performance plateaus.

\subsection{RQ3: Can Model Confidence Guide Selective Automation?}

We first evaluate the entity-specific classifiers trained with explicit negative sampling. Table~\ref{tab:ner_performance} summarizes performance on a unified 511-sentence test set created for the automation experiments. All four entity types show lower performance here than in Table~\ref{tab:ner-gold} because RQ1 evaluates only on sentences that contain the target entity, whereas the automation setting evaluates on the full test set, including sentences where the entity is absent. This introduces explicit negative examples and reflects the real-world scenario in which the model must detect the presence of an entity and avoid false positives. This more realistic evaluation setting naturally reduces F1 across all categories, with uncertain observations (OBS-U) being the most affected. ANAT-DP and OBS-DA achieve strong F1 scores (0.91 and 0.93), OBS-DP shows moderate reliability (0.83), and OBS-U remains challenging, with F1 of 0.62, reflecting the inherent ambiguity of uncertain findings. These models form the basis for the confidence-based automation analysis.

\begin{table*}[!htbp]
\centering
\caption{Annotation performance with negative sampling (per-label results on the 511-sentence test set)}
\label{tab:ner_performance}
\begin{tabular}{lccccc}
\toprule
\textbf{Entity Type} & \textbf{Positive Count} & \textbf{Negative Count} & \textbf{Precision} & \textbf{Recall} & \textbf{F1} \\
\midrule
ANAT-DP & 2033 & 392  & 0.9060 & 0.9172 & 0.9116 \\
OBS-DP  & 1856 & 569 & 0.9084 & 0.7669 & 0.8317 \\
OBS-DA  &  552 & 1500 & 0.9191 & 0.9381 & 0.9285 \\
OBS-U   &  309 & 1300 & 0.4930 & 0.8235 & 0.6167 \\
\bottomrule
\end{tabular}

\end{table*}

To determine confidence thresholds for automation, we split the 511-sentence test set evenly into 256 validation set and 255 held-out test set. Using the threshold discovery method described in Methodology section, Table~\ref{tab:threshold-summary} shows the minimum confidence thresholds achieving target entity match scores on the validation set. Threshold behavior varies substantially across categories: ANAT-DP and OBS-DA reach high match scores at modest thresholds, OBS-DP improves only under more aggressive filtering, and OBS-U never attains acceptable match scores. Because OBS-U does not reach any target match score, we route all reports containing an OBS-U prediction to manual review.
\begin{table}[htbp]
\centering
\caption{Minimum Confidence Thresholds Required to Reach Target Match Score Levels (Validation Split)}
\label{tab:threshold-summary}
\begin{tabular}{l c c c c}
\toprule
\textbf{Entity Type} & \textbf{0.80 Score} & \textbf{0.85 Score} & \textbf{0.90 Score} & \textbf{0.95 Score} \\
\midrule
ANAT-DP & 0.00 & 0.00 & 0.74 & -- \\
OBS-DP  & 0.36 & 0.91 & --   & -- \\
OBS-DA  & 0.00 & 0.79 & 0.95 & 0.97 \\
OBS-U   & --   & --   & --   & -- \\
\bottomrule
\end{tabular}

\end{table}

To avoid relying on match score alone, we report precision and recall at coverage. For ANAT-DP, increasing the threshold from 0.00 to 0.74 reduces coverage from 0.996 to 0.874 while improving precision from 0.903 to 0.941 (recall 0.926). For OBS-DA, increasing the threshold from 0.00 to 0.95 reduces coverage from 1.000 to 0.890 while improving precision from 0.899 to 0.929 (recall 0.981). For OBS-DP, increasing the threshold from 0.36 to 0.91 reduces coverage from 0.988 to 0.646 while improving precision from 0.927 to 0.978 (recall 0.764). We then apply the learned thresholds to full reports in the held-out test set using 90\% and 95\% entity-level acceptance rates (Table 6(a) and 6(b)). At the most permissive thresholds, 90\% of reports are automated with 0.86-0.87 match score. Stricter thresholds reduce coverage to between 54 to 63\% but increase automated match score to 0.91 to 0.92. Reports routed for manual review consistently show lower scores, indicating that confidence-based filtering successfully isolates difficult cases. Finally, under best-performing strict configuration, with 95\% acceptance, the system automates 140 of 255 reports, reducing manual annotation time from 510 to 230 mins (2 min/report), a 55\% reduction in human effort.
\begin{table}[ht]
\centering
\caption{Final Automation Results (a) 90\% (top) and (b) 95\% (bottom) Cutoffs}
\label{tab:final-automation}

\begin{subtable}{0.95\linewidth}
\centering
\label{tab:final-automation-90}
\begin{tabular}{c c c c c c c c}
\toprule
ANAT-DP & OBS-DP & OBS-DA & OBS-U &
Accept Count & Review Count &
Accept Score & Review Score \\
\midrule
0.00 & 0.36 & 0.00 & 0.00 & 229 & 26  & 0.865 & 0.712 \\
0.00 & 0.91 & 0.79 & 0.00 & 161 & 94  & 0.895 & 0.771 \\
0.74 & 0.91 & 0.95 & 0.00 & 141 & 114 & 0.916 & 0.774 \\
0.74 & 0.91 & 0.97 & 0.00 & 138 & 117 & 0.917 & 0.777 \\
\bottomrule
\end{tabular}
\vspace{0.3em}

\end{subtable}

\vspace{0.8em}

\begin{subtable}{0.95\linewidth}
\centering
\label{tab:final-automation-95}
\begin{tabular}{c c c c c c c c}
\toprule
ANAT-DP & OBS-DP & OBS-DA & OBS-U &
Accept Count & Review Count &
Accept Score & Review Score \\
\midrule
0.00 & 0.36 & 0.00 & 0.00 & 228 & 27   & 0.867 & 0.706 \\
0.00 & 0.91 & 0.79 & 0.00 & 160 & 95  & 0.897 & 0.769 \\
0.74 & 0.91 & 0.95 & 0.00 & 140 & 115  & 0.919 & 0.772 \\
0.74 & 0.91 & 0.97 & 0.00 & 137 & 118  & 0.919 & 0.775 \\
\bottomrule
\end{tabular}
\vspace{0.3em}

\end{subtable}

\end{table}

\textit{Evaluation:} Tables 6(a) and 6(b) summarize the
accept/review counts, and entity match score for accepted vs. review sets; rows correspond to threshold combinations for target match levels (0.80, 0.85, 0.90, 0.95).
As a sensitivity check for LLM overconfidence, we applied post-hoc isotonic calibration (3-fold cross-validation) and evaluated using Expected Calibration Error (ECE). 
That improved probability alignment across entities (OBS-DP ECE 0.0679→0.0341; OBS-U ECE 0.3468→0.1995). With thresholds at the 0.80 target match level, calibration reduced automatic acceptance by ~20\% while increasing automated match score from 0.865 (Table 6(a)) to 0.890, reflecting more conservative confidence estimates. The stricter targets showed similar behavior, but OBS-U did not reach  target levels.

\section{Discussion and Limitations}


Our findings show that entity-level performance is driven by both category frequency and linguistic consistency. Synthetic data proved effective in low-resource settings: with only 50 gold reports, augmentation with synthetic reports increased OBS-U's F1 from 0.61 to 0.70. However, as gold supervision increases, the marginal value of synthetic examples diminishes. Confidence-based routing effectively isolated reliable predictions, enabling a selective automation workflow. For a 10,000-report corpus under a 95\% acceptance rule, RadAnnotate could automatically process 55\% of cases, reducing expert effort from 333 hours to approximately 150 hours at an estimated 2 minutes per report. Limitations of this work include the need for further validation across diverse modalities and reporting styles, as the RadGraph test split provides only an initial cross-institution evaluation via CheXpert. This study is restricted to entity extraction, and excluding relation extraction limits full RadGraph compatibility. Furthermore, synthetic data may inherit source-corpus biases, and confidence thresholds learned from internal data likely require new environment recalibration. Finally, the effect of selective automation on downstream clinical utility remains unassessed.

\section{Conclusion}

RadAnnotate demonstrates that lightweight, entity-specific LLM classifiers can provide reliable RadGraph-style annotations even with modest amounts of gold data. Synthetic reports offer clear value when supervision is scarce, particularly for uncertain observations, and enable performance gains in low-resource settings. Confidence-based routing further translates model predictions into a practical workflow by automatically handling stable cases while reserving ambiguous findings for expert review. When deployed at scale, these components together can reduce radiologist annotation time 50-60\% for large backlogs of new reports. Thus, targeted LLM pipelines with selective automation can make institutional-grade clinical NLP dataset creation significantly more feasible.

\makeatletter
\renewcommand\@biblabel[1]{#1.}
\makeatother
\makeatletter
\renewcommand{\bibsection}{%
  \begin{center}
    \normalfont\bfseries References
  \end{center}}
\makeatother
\bibliographystyle{vancouver}
\bibliography{references}

\end{document}